\documentclass[sigconf]{acmart}
\AtBeginDocument{%
  \providecommand\BibTeX{{%
    \normalfont B\kern-0.5em{\scshape i\kern-0.25em b}\kern-0.8em\TeX}}}

\usepackage{breakcites}
\usepackage{microtype}
\usepackage{graphicx}
\usepackage{subfigure}
\usepackage{booktabs} % for professional tables

\usepackage{amssymb,amsmath}
\usepackage{pifont}
\usepackage{color}
\usepackage[english]{babel}
\usepackage{array}
\usepackage{bm}
\usepackage{multirow}
\usepackage{enumitem}
\usepackage{pifont}
\usepackage{tikz}
\usepackage{amssymb}
\usepackage{enumitem}
\usepackage[english]{babel}
\usepackage{array}
\usepackage{bm}
\usepackage{multirow}
\usepackage{enumitem}
\usepackage{pifont}
\usepackage{graphicx}
\newcommand{\eat}[1]{}
\usepackage{flushend}
\usepackage{makecell}
\usepackage{pifont}
\acmConference[DAC]{DAC}{Design Automation Conference}{2022}
\begin{document}

\title{The Larger The Fairer? Small Neural Networks Can Achieve Fairness for Edge Devices}

% \IEEEauthorblockN{ Yi Sheng$^{1}$ \quad Junhuan Yang$^{2}$ \quad Yawen Wu $^{3}$ \quad Kevin Mao $^{4}$ \quad Yiyu Shi $^{5}$ \quad Jintong Hu $^{3}$ \quad Weiwen Jiang$^{1,*\thanks{* Weiwen Jiang is the corresponding author (wjiang2@nd.edu)}}$ \quad Lei Yang $^{2}$}  
\author{Yi Sheng}
\affiliation{%
  \institution{George Mason University}
}
\author{Junhuan Yang}
\affiliation{%
  \institution{The University of New Mexico}
}
\author{Yawen Wu}
\affiliation{%
  \institution{University of Pittsburgh}
}
\author{Kevin Mao}
\affiliation{%
  \institution{University of Pittsburgh}
}
\author{Yiyu Shi}
\affiliation{%
  \institution{University of Notre Dame}
}
\author{Jingtong Hu}
\affiliation{%
  \institution{University of Pittsburgh}
}
\author{Weiwen Jiang}
\affiliation{%
  \institution{George Mason University}
}

\author{Lei Yang}
\affiliation{%
  \institution{The University of New Mexico}
}

% \affiliation{%

\renewcommand{\shortauthors}{}

\begin{abstract}
% With the breakthrough of AI democratization, automated AI has been increasingly deployed in edge and mobile devices to provide healthcare, from mobile dermatology assistant, mobile eye cancer detection, to comprehensive vital signs monitoring. 
% While these techniques rely on visual assistance like the cameras that come with mobile devices and inevitably lead to different levels of fairness concerns, in addition to the inherently high requirements on accuracy and hardware efficacy.
% Neural architecture search (NAS) approaches have been proposed as an effective way for exploring optimized neural network which can be accommodated on these devices with a better accuracy-performance trade-off; however, the fairness issues, such as the racial, skin color, and socioeconomic inequities, are ignored, leading to the increasing healthcare disparities. 
% In this work, we propose a hardware and software co-exploration framework,  leveraging NAS for achieving neural networks for fairness on edge devices for skin disease.
% We can not only shrink the neural network model for edge and mobile devices, but also we can improve the model to be fairer by incorporating more equally represented data.

Along with the progress of AI democratization, neural networks are being deployed more frequently in edge devices for a wide range of applications. Fairness concerns gradually emerge in many applications, such as face recognition and mobile medical. One fundamental question arises: what will be the fairest neural architecture for edge devices? By examining the existing neural networks, we observe that larger networks typically are fairer. But, edge devices call for smaller neural architectures to meet hardware specifications. To address this challenge, this work proposes a novel \underline{Fa}irness- and \underline{Ha}rdware-aware \underline{N}eural \underline{a}rchitecture search framework, namely FaHaNa.
Coupled with a model freezing approach, FaHaNa can efficiently search for neural networks with balanced fairness and accuracy, while guaranteed to meet hardware specifications.
% , latency, and accuracy. 
Results show that FaHaNa can identify a series of neural networks with higher fairness and accuracy on a dermatology dataset. Target edge devices, FaHaNa finds a neural architecture with  slightly higher accuracy, 5.28$\times$ smaller size, 
%3.52\%
{15.14\%} higher fairness score, compared with MobileNetV2; meanwhile, on Raspberry PI and Odroid XU-4, it achieves 5.75$\times$ and 5.79$\times$ speedup.

% a neural network with 5.28$X$ smaller size  and 

% with the highest fairness 

% we can identify neural architectures with higher accuracy and fairness but halve the model size for edge devices.

\end{abstract}

% % https://www.overleaf.com/project/6170ce1a3ac2ca732268e7a3


% \keywords{datasets, neural networks, gaze detection, text tagging}

\maketitle

\vspace{-5pt}
\section{Introduction} \label{sec:Intro}
With the continuous progress of AI democratization, we have witnessed the breakthrough of deep learning models deployed in the edge and mobile devices for AI applications, like mobile dermatology assistant~\cite{googleai}, mobile eye cancer detection~\cite{cradle}, comprehensive vital signs monitoring~\cite{binah}, and medical imaging and diagnostics~\cite{kaissis2020secure}.
%are benefited to provide healthcare in our life. , emotion detection~\cite{emotionai},
To implement these models efficiently on devices, various model compression, accelerator design, and hardware/software co-design techniques \cite{han2015deep,zhang2015optimizing,hao2019fpga,song2021dancing,peng2021optimizing, jiang2020device, jiang2019achieving, zhang2019neural, jiang2019accuracy} have been proposed to achieve both high accuracy and efficiency. 
%These breakthroughs, pruning, xxx, design, provide neural networks with high accuracy for tasks on different datasets and the efficient hardware implementations to run these neural networks. 
Unfortunately, most of the existing AI system designs only pursue high overall accuracy and ignore fairness among diverse groups in the dataset. For example, \cite{biasapp} has pointed out the gender and skin-type bias in commercial AI systems. Examination of facial-analysis software shows an error rate of 0.8\% for light-skinned men, 34.7\% for dark-skinned women; 
\cite{kamulegeya2019using} also pointed out similar racial disparity for Skin Image Search, which is an AI app that helps people identify skin conditions. It reports 70\% accuracy for the whole dataset, but only 17\% for dark skins.

%All of these techniques rely on visual assistance, such as cameras, that come with mobile devices and inevitably lead to different levels of fairness concerns. 
%One of the main reasons that contribute to this disparity is that existing easily accessible data sets are inherently biased.

%In a wide range of AI applications, fairness is a core requirement.

\begin{figure}[t]
\vspace{-3pt}
  \centering
  \includegraphics[width=3.2816 in]{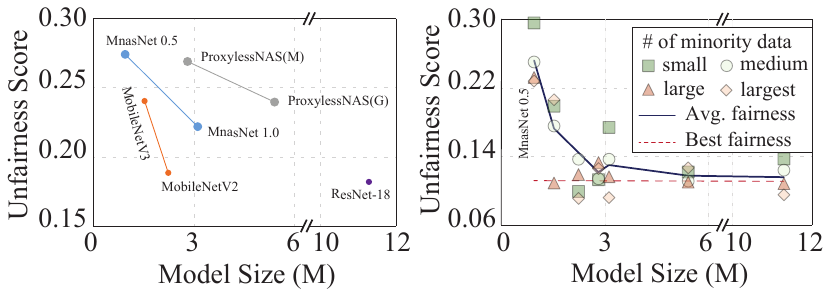}
  \vspace{-3pt}
  \caption{Fairness vs. model size on the existing neural networks: (a) larger networks within the same series have higher fairness; (b) increasing fairness along with larger models.}
  \label{fig:intro}
  \vspace{-5pt}
\end{figure}

%1. system
%2. manual 
Research efforts have been made in addressing the fairness issue~\cite{mehrabi2021survey}.
However, they either focus on the model interpretability by modifying the neural network models to be fairer \cite{choras2020machine}, or fairness-aware data collection \cite{choi2020fair}.
While these works make important initial steps, achieving fairness on resource-constrained edge devices brings new challenges: neural networks need to be small enough to accommodate limited computation power and memory/storage space.
However, as shown in Figure \ref{fig:intro}, we observed that larger neural network models generally have higher fairness, where the ``\textbf{unfairness {score}}'' is defined as the variation of the prediction accuracy among the diverse groups. Thereby, a fundamental question we are trying to answer is: can we identify small and fair neural networks to meet the hardware specifications? What is more, traditional methods manually fine-tune the models to achieve better fairness. In this work, we are trying to achieve fairness through automatic neural architecture search (NAS).

%Research works have shown that AI fairness can be improved by algorithm-level optimization, such as datasets balancing~\cite{choi2020fair};
% \cite{buolamwini2018gender,jo2020lessons,9521312,he2020geometric,oneto2020fairness};
% and the model training \todo{cite} and \todo{maybe others}
%yet, there is still a vacancy to achieve AI fairness from system-level optimization.

%When it comes to the design of AI systems, in particular for edge computing, the limited storage space and computation power bring the biggest challenge: neural networks need to be small enough to be accommodated on these devices.

% Towards this, a novel ``Fairness- and Hardware-aware NAS'' framework, namely FaHaNa, is proposed in this paper. 
% It integrates the fairness as a part of the reward in a reinforcement learning based optimization process.
% To the best of our knowledge, this is the first work to bring 

%In the AI democratization era, we have witnessed the breakthroughs of AI system automation in the past few years: from the automation of neural network design (i.e., neural architecture search, NAS)~\cite{zoph2016neural} to hardware-aware NAS~\cite{tan2019mnasnet,cai2018proxylessnas} and hardware/software co-design NAS~\cite{hao2019fpga,jiang2019accuracy}.

Although there have been various NAS frameworks~\cite{zoph2016neural,tan2019mnasnet,cai2018proxylessnas,zhang2021dian, yan2020ms, bian2020nass, jiang2020hardware, yang2020co, yang2020co-ex, lu2019neural}, none of them have considered fairness as a goal. In this paper, we propose a novel ``Fairness- and Hardware-aware NAS'' framework, namely FaHaNa, to address these challenges. 
It integrates fairness as a part of the objective in a reinforcement learning (RL) based optimization process.
Given a target hardware platform and a training dataset with diverse groups, FaHaNa searches for the neural architectures with the highest accuracy and the best fairness. Meanwhile, the latency can be guaranteed to meet the hardware specifications.
To ensure fairness awareness, it seems straightforward to simply include a fairness metric together with accuracy to the existing NAS frameworks. 
However, this extra fairness metric can easily pull down good candidates (in terms of accuracy) in the search space, since they usually achieve high accuracy by catering to the majority group. Therefore, the NAS framework needs to ensure high fairness for diverse groups, while not compromising overall accuracy.
%As such, the discrimination among models becomes vague, which will potentially extend the search time.
In addition, NAS itself is known for lengthy search times.
% Problems like those challenge and haunt the efficacy of the design as well as the optimization of the fair neural networks.
%On all these counts, a more efficient approach is highly demanded to conduct the fairness-aware neural architecture exploration.

% Fairness-aware neural architecture exploration provides a neat shortcut to addressing above issues.

To address these challenges, our proposed NAS framework leverages a dedicated designed reward function to balance fairness, accuracy, and hardware efficiency.
Furthermore, we observed that the front layers (header) of neural networks will not affect fairness but only extract common features; while the intermediate feature maps in the end layers (tail) are quite different.
Based on this observation, we develop a freezing method to accelerate the optimization without affecting the fairness.
% some layers in the network 
% the majority and minority groups have similar intermediate features in the front layers (header), while the features in the end layers (tailor) are quite different.
%we novelly put forward a freezing method to accelerate the optimization, where a part of the backbone neural architecture can be fixed with the pre-trained parameters (weights).
As a result, the training parameters and training time can be reduced, together with the reduction in the search space.
% The freezing method is 
% % for 
% % both the training time and the search space can be reduced.
% % can be applied without training during NAS,
% % , which is 
% motivated by our observation that, in existing pretrained neural networks, the majority and minority groups have similar intermediate features in the front layers (header), while the features in the end layers (tailor) are quite different.
% As such, we can freeze the header and only search for the tailor.
% % to freezing the header (i.e., front layers) of the backbone neural architecture, such that the pre-trained models can be used in the header without training during NAS, consequently both the training time and the search space can be reduced.
% To enable this, we devise an algorithm to identify the header to extract the common features.
The main contributions of this paper are as follows.
\begin{itemize}
  \item \textbf{Framework.} To the best of our knowledge, FaHaNa is the first fairness-aware framework to explore fair neural architectures, which can further generate the optimal DNN architectures with the guaranteed latency on target hardware. 
  \item \textbf{Acceleration.} We propose a freezing method to fix a part of the neural architecture and make use of the pre-trained parameters for common feature extraction, which significantly improves search efficiency without affecting the fairness.
  \item \textbf{Evaluation.} We have conducted a case study on medical AI (i.e., dermatological disease diagnosis) to evaluate FaHaNa. A dermatology dataset, including images with light skin (majority) and dark skin (minority), is built for evaluation.
\end{itemize}

{Experimental results on the dermatology dataset evaluate the effectiveness of FaHaNa and the efficiency of the freezing method to accelerate the optimization process.
First, compared with MnasNet, the network identified by FaHaNa (FaHaNa-Nets) can reduce the unfairness score from 0.4521 to 0.1973, meanwhile achieving 3.16\% overall accuracy gain, 2.24$\times$ smaller model size, 2.11$\times$ and 3.15 $\times$ latency reduction on Raspberry PI and Odroid XU-4. 
Compared with a larger but fair model, MobileNetV2, FaHaNa-Nets can achieve 15.14\% higher fairness and 0.23\% higher accuracy, and the reductions of model size and latency are increased to 5.28$\times$, 5.75$\times$, and 5.79$\times$.
Second, the freezing method is effective to better explore the design space, reducing the search space from $10^{19}$ to $10^{9}$ and accelerates the search process with 2.67$\times$ speedup.
Last but not the least, FaHaNa is compatible with existing fairness techniques \cite{choi2020fair}.
% Results show that FaHaNa-Nets can consistently achieve higher fairness with 
}

% common data set such as ImageNet show that in the cases where the state-of-the-art \cite{zoph2016neural} generates architectures with latencies $7.81\times$ longer than the specifications, those from FNAS can meet them with less than $1\%$ accuracy loss; meanwhile FNAS also achieves $11.13\times$ speedup for the search process. 

In the rest of the paper: Section 2 reviews the related background and provides the motivations; Section 3 defines the problem and presents our FaHaNa framework. Experimental results are shown in Section 4 and concluding remarks are given in Section 5.

\section{Related Work and Motivation}\label{sec:pre}

% \todo{\noindent\textbf{A. Related work}}

% \noindent\textbf{B. Observations and motivations}
This section will provide our observations on the effects of neural architectures on fairness and review the related works. 

{\textbf{Observation 1:} Neural architectures affect fairness.}

\begin{figure}[t]
%\vspace{-10pt}
  \centering
  \includegraphics[width=3.1 in]{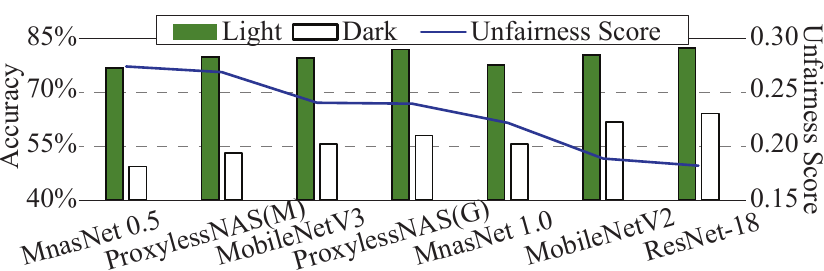}
%   \vspace{-10pt}
  \caption{Neural architectures affect fairness.}
  \label{fig:mot1}
  \vspace{-8pt}
\end{figure}

On the dermatology dataset, Figure \ref{fig:mot1} shows the unfairness {score} of different sets of neural architectures, including MobileNet, MnasNet, ProxylessNAS, and ResNet.
The green bars and white bars represent the prediction accuracy of the majority (light skins) and minority (dark skins) in the dataset, respectively.
The blue line shows the unfairness {score} %metrics 
on all models, which describes the variance in accuracy between the majority and minority groups.
% the minority suffers a bias
% are used to evaluate the fairness. The pictures in dataset ISIC2019 are almost for white people. To evaluate fairness, the dataset Altas, Derm, and Black which contain pictures of other color people are used together to train the neural network models. Specifically, we use the result for the dataset group which contains ISIC2019 and 200 pictures from the 3 datasets to show the effect on fairness. As shown in Figure x, different neural architectures affect fairness a lot. The line in the figure shows the accuracy gap between white people and other color people. 
% According to the result, 
More specifically, the unfairness {score} varies from {0.4521}
%27.42\% 
(MnasNet 0.5) to {0.1820}
%18.20\% 
(ResNet-18) 
as reported in the figure.
Results demonstrated that all these models have prejudice on the majority models, and each model has better fairness than its left-hand ones.

{\textbf{Motivation 1}: Searching for a fair neural architecture.}

The straightforward and commonly applied approach to address fairness issue is to balance data between the majority and minority groups \cite{choi2020fair}
% \cite{buolamwini2018gender,jo2020lessons,9521312}, 
or learn fair representations between the protected and unprotected features~\cite{he2020geometric}.
However, there exists an inherent imbalance since data from the minority groups may not be easily collected due to objective reasons (e.g., a lack of medical professionals from marginalized communities). 
What's worse, neural architecture acts as an equal or even more important role in fairness, and the effects of different network models may outweigh that by data balancing.
% increase the samples from the minority groups to   \cite{buolamwini2018gender,jo2020lessons}.
% Related works: Existing approach proposed to resolve the fairness issue by proposing strategies for data collecting~\cite{buolamwini2018gender,jo2020lessons}, or strait forwardly increasing the number of inputs for the minority groups~\todo{\cite{}}. However, we have observed that simply increasing minority data may not be enough.}
% a better neural architecture may achieve better fairness even with a small number of than others.
% increasing minority data only may not pro}
% \todo{----Balance the data is not an effective way for increasing fairness, the accuracy could not be correspondly increased.}
Results in Figure~\ref{fig:intro}(b) show the unfairness of different neural architectures on the training datasets with different amounts of minority data.
We observe that even MnasNet 0.5 is trained on a dataset \textbf{with $\bf 5\times$ minority data} (i.e., diamond for the smallest model), its unfairness score is still higher than ResNet-18 ({0.2280}
%22.80\% 
vs.
{0.1820}).
%18.20\%).
This emphasizes the effects of the neural architecture on fairness and motivates us to conduct the fairness-aware architecture search.

% increase the minority data to ,  still has higher unfairness score than ResNet-18 (22.80\% vs. )

% always have higher unfairness score

%  the 

% There are a lot of neural architectures which are suitable for different scenarios. However, for the fairness-demanded situation, it is not easy to identify fair neural architecture. After doing a lot of experiments, we may be able to distinguish the fair architecture. Nevertheless, it is costly to do so, and even sometimes it is not easy to find such architecture. So, can we find fair architecture without experiments in different situations? It is a great motivation for us to find an easy way to identify fair architecture.

% \subsection{Motivation 2}
\vspace{3pt}
{\textbf{Observation 2:} Hardware specification affects fairness.}

Table \ref{tab:mot2} reports the accuracy, unfairness {score}, and hardware performance of different neural network models.
We run these models on Raspberry PI with a timing constraint of 1500ms.
With such a hardware constraint, only SqueezeNet 1.0, MobileNetV3, and MnasNet 0.5 can meet the specification; however, the unfairness scores of MnasNet 0.5 and MobileNetV3 are {0.2196}
%21.96\% 
and {0.0928}
%9.28\%
less than MobileNetV2's score. Nevertheless, its latency violates the requirement.
% \todo{The figure for MobileNetV3 is larger than 5\%.}
SqueezeNet 1.0 is much fairer, but its accuracy is as low as 15.65\%.
These results clearly demonstrate that fairness cannot be considered separately from hardware specifications.
\begin{table}[t]
%\vspace{-5pt}
  \centering
  \small
  \tabcolsep 3.5pt
  \renewcommand\arraystretch{1.0}
  \caption{Different models with less than 30MB storage size running on Raspberry PI with timing constraint of 1500ms}
    \begin{tabular}{cccccc}
    \hline
    \multirow{2}{*}{\textbf{Model}} & \textbf{Latency} & \textbf{Storage} & \multirow{2}{*}{\textbf{Accuracy}} & {\textbf{Unfairness}} & \textbf{Meet} \\
    \textbf{} & \textbf{(ms)} & \textbf{(MB)} & \textbf{} & \textbf{Score} & \textbf{Spec.} \\
    \hline
    
    SqueezeNet 1.0 & 122.92 & 2.77  & 15.65\% & 0.2159 & \checkmark \\
    MobileNetV3 & 658.84 & 5.81  & 80.38\% & 0.3253 & \checkmark \\
    MnasNet 0.5 & 714.19 & 3.60  & 78.12\% & 0.4521 & \checkmark \\
    \hline
    MobileNetV2 & 1,939.40 & 8.51  & 81.05\% & 0.2325 & $\times$ \\
    ProxylessNAS(G) & 3714.44 & 20.60  & 83.21\% & 0.2667 & $\times$ \\
    MnasNet 1.0 & 3855.72 & 11.86  & 80.71\% & 0.2913 & $\times$ \\
    ProxylessNAS(M) & 5241.51 & 10.70  & 81.27\% & 0.3094 & $\times$ \\
    % ResNet-34 & 621.87 & 81.20  & 83.01\% & 23.97\% & $\times$ \\
    % ResNet-50 & \multicolumn{1}{c}{1063.61} & \multicolumn{1}{c}{89.72 } & 83.81\% & 18.55\% & $\times$ \\

    % SqueezeNet 1.0 & 122.92 & 2.77 & 15.65\% & 21.59\% & \checkmark \\
    % MobileNetV3 & 658.84 & 5.81  & 79.39\% & 24.05\% & \checkmark \\
    % Mnasnet 0.5 & 1159.08 & 3.60  & 76.55\% & 27.42\% & \checkmark \\
    % \hline
    % MobileNetV2 & 1939.40 & 8.51  & 80.41\% & 18.86\% & $\times$ \\
    % ProxylessNAS(G) & 3714.44 & 20.60  & 81.77\% & 23.97\% & $\times$ \\
    % ProxylessNAS(M) & 5241.51 & 10.70  & 79.74\% & 26.90\% & $\times$ \\
    % MnasNet 1.0 & 7793.18 & 11.86   & 77.54\% & 22.19\% & $\times$ \\
    % Resnet-34 &  &   &  &  & \\
    % Resnet-50 &  &  &  &  & \\
    \hline
    \end{tabular}%
  \label{tab:mot2}%
\end{table}%

%\vspace{3pt}
{\textbf{Motivation 2:}} Making tradeoffs among fairness, accuracy, and hardware efficiency

% The neural architecture needs not only provide high fairness but also satisfy the hardware specifications}

% To identify a fair neural network to meet hardware specifications,

%\todo{----(AutoML.org --->Multi-objective NAS), search some related paper form high-quality conferences, latency-aware, power, performance, while they only consdier either, without fairness} 

Fairness, accuracy, and hardware efficiency are equally important in edge AI applications, like medical AI~\cite{kaissis2020secure,wu2021medical}.
Losing any one of these characteristics will render the architecture useless (e.g., SqueezeNet has low accuracy, MobileNetV2 violates latency, and MnasNet 0.5 is less fair). Holistic optimization should be conducted on all these metrics.

Neural architecture search (NAS) methods have been developed to automatically identify neural architectures for maximum accuracy~\cite{zoph2016neural}. Together with the consideration of the hardware specifications, hardware-aware NAS~\cite{tan2019mnasnet,cai2018proxylessnas, jiang2020standing} further explore the hardware design space, thus jointly identifying the best architecture and hardware designs.
Decoupled from hardware, the multi-objective NAS (MONAS)~\cite{hsu2018monas} was proposed.
% recently framed by taking, e.g., resource requirements into account, via co-optimization of performance with other criteria like error resilience~\cite{Schorn2020AutomatedDO}, scalability~\cite{awad2021dehb}, robustness~\cite{zimmer2021auto}, and etc.
Nevertheless, there is still a lack of NAS considering the fairness in the design objective.
Straightforwardly integrating fairness into MONAS will reduce the reward of models with high accuracy but low fairness, and make the discrimination among models to be vague.
As such, it potentially prolongs the search process for convergence.
% brings challenges for identifying the optimal solutions.
Furthermore, NAS itself is known for its lengthy search time.
Therefore, a more efficient way for fairness-aware NAS is highly demanded.

\begin{figure}[t]
%\vspace{-10pt}
  \centering
  \includegraphics[width=3.1 in]{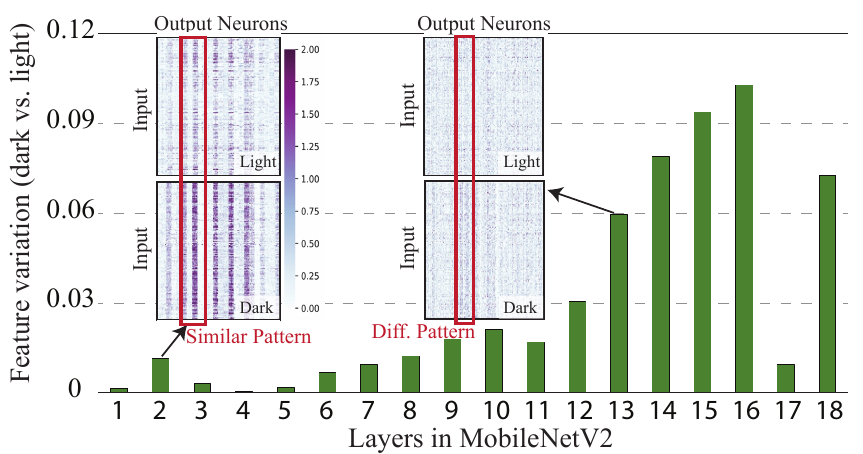}
%   \vspace{-10pt}
  \caption{Header extracts common features and fairness is mostly affected by tail in a neural network. Note that layer 17 has small variation since most elements approach to 0.}
  \label{fig:mot3}
  %\vspace{-8pt}
\end{figure}

\textbf{Observation 3:} Fairness is mostly affected by the tail.

% No matter the network is fair or not, minority and majority have similar results in first layers.

To figure out how to accelerate the NAS process, we further investigated: how do the neural networks make different predictions for the minority or majority groups? 
Toward this, we compare the variation of intermediate features obtained by different groups after each layer in MobileNetV2.
Results in Figure \ref{fig:mot3} show that the front layers (say before layer 12) have small variations.
The visualizations of features after layer 2 and layer 13 are illustrated in Figure \ref{fig:mot3}, where each row represents the intermediate feature corresponding to one input data, and the column corresponds to a specific neuron.
Visualized pictures show that layer 2 has small variation because it has similar patterns in features from different groups, while layer 13 has different patterns.
More sets of experiments on other networks have been conducted, and we obtain the same observation.

% indicates the output neuron.

% , and layer 17 are shown. For layer 2, there are some obvious similar patterns shown in the visualization result of white people picture and other color people picture. For layer 13, the patterns in the two visualization results become less obvious. For layer 17, it seems that the two visualization results are very similar and there are no patterns.

% plots the variation of intermediate features from majority inputs and minority inputs after each layer in MobileNetV2. The results show that 

% We use one picture from the minority dataset and majority dataset to do inference to explore this interesting question. Figure \ref{fig:motivation3} shows the visualization result of the neural network models MobileNetV2 trained by our datasets. The bar chart shows the difference between the picture for white people and the picture for other color people.  As the figure is shown, in the first 8 layers, the result of a picture from the 3 small datasets has similar characteristics to that of the picture from the big dataset. To illustrate the phenomenon better, the visualization results of layer 2, layer 13, and layer 17 are shown. For layer 2, there are some obvious similar patterns shown in the visualization result of white people picture and other color people picture. For layer 13, the patterns in the two visualization results become less obvious. For layer 17, it seems that the two visualization results are very similar and there are no patterns.

\textbf{Motivation 3:} Freezing the head and searching for the tail.

The above results demonstrate that the distinctions of groups are mainly contributed by the end layers; in other words, the front layer(s) extracts the common features which will not affect fairness. 
Based on the observation, we are inspired to freeze the header in the search process, and only search for the architecture of the tail.

\section{FaHaNa: Put Fairness, Hardware, NAS in a Holistic Optimization Loop}

\noindent\textbf{3.1 Problem Definition}

%AI fairness has different definitions in specific machine learning tasks.
%This paper studies the fairness issue targeting classification task in computer vision.
In this work, we study the fairness issue on the classification task in computer vision.
This section will formally define the problem of ``fairness-hardware-neural-architecture co-optimization''.

% co-exploration of neural networks on target hardware architecture.

\textbf{Classification.} Given a dataset $D$, we define $C=\{c_1,c_2,\cdots,c_M\}$ as a set of $M$ classes, where each data $d_i\in D$ belongs to a class $c_j\in C$. That is, there exists a mapping function $f$: $f(d_i)=c_j$.
A neural network $N$ is to build the mapping function from $D$ to $C$.
On top of a training dataset, $N$ will learn a function $f_N^{\prime}$ to approximate $f$.
% A machine learning model (or classifier in classification task), denoted as $f^{\prime}$, is to approximate $f$.
If $f(d_i)=f_N^{\prime}(d_i)$, it is a correct prediction on data $d_i$; otherwise, it is an incorrect prediction.
The accuracy $A(f_N^{\prime},D)$ describes the ratio of data in $D$ getting the correct prediction using model $N$.

\textbf{Diverse Groups.} For each data $d_i\in D$, in addition to its category feature (i.e., $C$), it may also have other inherent features, like the skin-color, race, sex, etc.
% These features can divide dataset $D$ into groups.
For an inherent feature $I$, it can divide $D$ into $K$ groups: $D=\{D_{g_1},D_{g_2},\cdots,D_{g_K}\}$.
Take the feature of skin color as an example, it can divide $D$ to 2 groups: light skin ($g_1=light$) and dark skin ($g_2=dark$).
% with $K$ groups: $G=\{g_1,g_2,\cdots,g_\}$,
% We define $G=\{g_1,g_2,\cdots,g_N\}$ to be an inherent feature for $N$ groups; say $G=\{g_1,g_2\}$ to represent skin-color feature with 2 groups: light skin ($g_1=light$) and dark skin ($g_2=dark$).
% Correspondingly, the dataset $D$ is divided into $N$ groups: $D=\{D_{g_1},D_{g_2},\cdots,D_{g_N}\}$.
If the number of data in $D_{g_i}$ is less than that in $D_{g_j}$, i.e., $|D_{g_i}|<|D_{g_j}|$,
% the groups has the relationship $|D_{g_1}|<|D_{g_2}|$, 
then we call $D_{g_i}$ (e.g., dark skin) minority group in comparison with $D_{g_j}$ (e.g., light skin).
Kindly note that the proposed method can support fairness for more than 2 diverse groups.

\textbf{Fairness.} For a model $N$ on data group $D_{g_k}$, its accuracy is $A(f_N^{\prime},D_{g_k})$.
Based on the accuracy of all groups, we define the unfairness score $U$ of a model $N$ on dataset $D$ based on L1-norm, which is $U(f_N^\prime,D)=\sum_{\forall g_i\in G}\{|A(f_N^{\prime},D_{g_i})-A(f_N^\prime,D)|\}$.

\textbf{Specification.} The specification contains two parts: software specification and hardware specification.
The software specification is the requirement for prediction accuracy. Given an accuracy constraint $AC$, it requires the model $N$ to achieve accuracy $A(f^\prime_N,D)\ge AC$.
As to the hardware specification, we will be given a hardware device $H$ (e.g., Raspberry PI, a mobile phone, etc.), with the timing constraint $TC$.
$L(H,N)$ represents the inference latency of running neural network $N$ on $H$.
The hardware specification $S$ sets up hardware performance requirements, such as $L(H,N)\leq TC$.
% and storage constraint $SC$.

\noindent\textbf{Problem Definition.} With the above definitions, we can formally define the ``fairness-hardware-neural-architecture co-optimization problem'' as follows: Given a dataset $D$ with $M$ classes and an inherent feature $I$ dividing $D$ into $K$ groups, a hardware $H$, design specifications (e.g., timing constraint $TC$ and accuracy constraint $AC$), our objective is to automatically generate a neural architecture $N$, such that the accuracy $A(f_N^{\prime},D)$ can be maximized and the unfairness score $U(f_N^\prime,D)$ can be minimized; meanwhile, accuracy $A(f_N^{\prime},D)$ and latency $L(H,N)$ can meet the design specifications.

\vspace{2pt}
\noindent\textbf{3.2 FaHaNa Framework}

\begin{figure}[t]
%\vspace{-10pt}
  \centering
  \includegraphics[width=3.1 in]{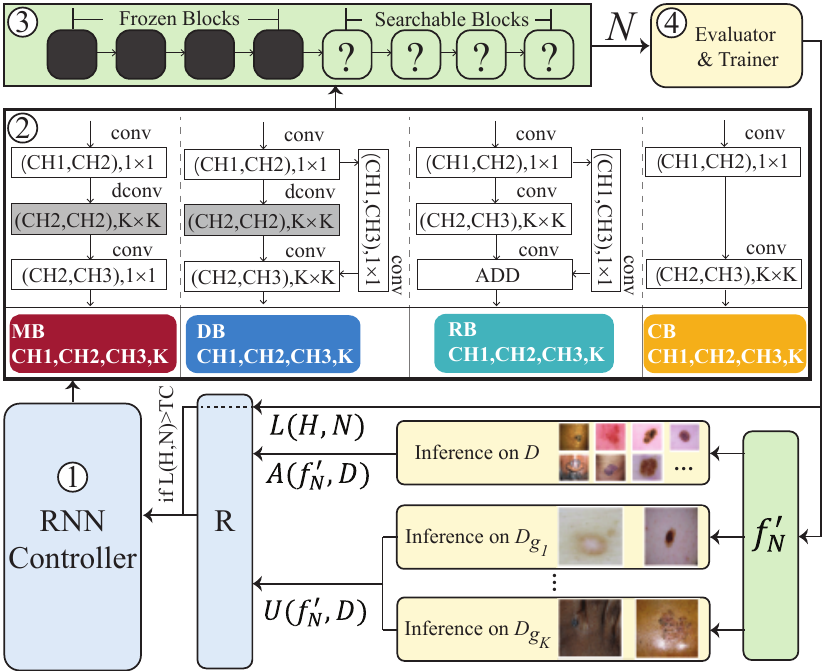}
%   \vspace{-10pt}
  \caption{Overview of FaHaNa framework.}
  \label{fig:framework}
  \vspace{-8pt}
\end{figure}

\textbf{FaHaNa Overview:} {Figure~\ref{fig:framework} illustrates the overview of our proposed FaHaNa framework. 
It is composed of four components: {\large\ding{192}} a recurrent neural network (RNN) based controller, {\large\ding{193}} a block-based search space, {\large\ding{194}} backbone architecture producer, {\large\ding{195}} performance evaluator and trainer.
Specifically, {\large\ding{192}} the controller will guide the optimization process. From {\large\ding{193}} block-based search space, it will identify the searchable block in the backbone architecture (obtained by {\large\ding{194}} producer) to form a neural network $N$ (a.k.a., child network).
Then, $N$ will be sent to the {\large\ding{195}} trainer to learn the function $f^\prime_N$.
It will be used for the inference on dataset $D$ and sub-group of $\{D_{g_1}, D_{g_2}, \cdots, D_{g_K}\}$ to obtain the accuracy $A(f_N^{\prime},D)$ and unfairness score $U(f_N^\prime,D)$, respectively.
Simultaneously, {\large\ding{195}} evaluator will get the latency $L(H,N)$ of $N$ on the given hardware $H$.
Finally, a reward will be generated to update RNN in the controller.
In the following section, we will introduce these components in detail.
}

% $A(f_N^{\prime},D)$ can be maximized, the unfairness score $U(f_N^\prime)$ can be minimized, meanwhile, the latency $L(H,N)$

% evaluator and trainer to obtain the latency $L(H,N)$ on hardware $H$, 

% freezing block and searchable block

% It will 

% first identify the neural architecture from the block-based search space.

% (2) Half NAS freezing accelerate,no training for freezing layers in order to speed up the search process. (3) Half NAS network choose,which you can choose different structure and connect it in the network (4) Performance evaluator,the most important part in the framework and it will give out the reward. We will introduce
% each component specifically as follows.}

\vspace{1pt}
{\large\ding{192}} \textbf{RNN Controller}: 
% \noindent\wwnote{{\bf(1) RNN based controller}}
The controller will iteratively predict the hyperparameters of a child network.
In each iteration (a.k.a., episode), the controller will receive a reward to update the RNN network.
The reward $R$ is generated based on the accuracy $A(f_N^{\prime},D)$, unfairness score $U(f_N^\prime,D)$, and latency $L(H,N)$ (details see {\large\ding{195}} Evaluator and Trainer), which is formulated as below.
\vspace{-6pt}
\begin{equation}\label{equ:reward}
\small
R = \left\{ {\begin{array}{*{20}{c}}
{\alpha\cdot A(f_N^{\prime},D)-\beta\cdot U(f_N^\prime,D)}&{L(H,N)\le TC,\  A(f_N^{\prime},D)\ge AC}\\
{-1}&{otherwise}
\end{array}} \right.
\end{equation}
where $\alpha$, $\beta$ are two scaling factors that could be adjusted according to the specific
% probably higher 
demands on accuracy or fairness.
Based on the reward, we employ reinforcement learning to update the controller.
Specifically, we apply Monte Carlo policy gradient algorithm \cite{williams1992simple}:
\vspace{-6pt}
\begin{equation}
\small
    \nabla J(\theta) = \frac{1}{m}\sum_{k=1}^{m}\sum_{t=1}^{T}\gamma^{T-t}\nabla_{\theta}\log \pi_{\theta} (a_{t}|a_{(t-1):1})(R_{k}-b)
\end{equation}
where $m$ is the batch size and $T$ is the number of steps in each episode. Rewards are discounted by
an exponential factor $\gamma$ and the baseline $b$ is the average exponential moving of rewards.

\vspace{1pt}
{\large\ding{193}} \textbf{Search Space}: As shown in Figure \ref{fig:framework}, the search space is based on different basic computation blocks.
Motivated by existing neural networks with the highest fairness (i.e., MobileNetV2 and ResNet-18 from Figure \ref{fig:mot1}), we consider 4 types of basic blocks (MB, DB, RB, and CB).
MB and DB are based on MobileNetV2 blocks with $stride=2$ and $stride=1$, respectively; RB is based on ResNet blocks; we also include CB based on the conventional convolution operation.
All these 4 blocks have the same hyperparameters: channel numbers ($CH1$, $CH2$, and $CH3$) and kernel sizes ($K$).
% As shown in Figure \ref{fig:framework} {\large\ding{193}}, each block has channel numbers: $CH1$, $CH2$, and $CH3$.
Kindly note that $CH1$ of one block $b$ is determined by $CH3$ of block $b$'s precedence, while $K$, $CH2$ and $CH3$ are searchable.
% In addition to channel number, the kernel size applied to each block is also searchable.
We also enable the skip operation in a block to make the flexibility on the depth of the neural network.

\vspace{1pt}
{\large\ding{194}} \textbf{Backbone Architecture Producer}: 
In the conventional NAS, each layer/block in the backbone architecture is searchable; however, motivated by \textbf{Observation 3} in Section \ref{sec:pre}, we develop a producer to freeze the header of the backbone architecture.
The challenge here is how to determine the blocks to be frozen or not, as shown in Figure \ref{fig:framework} {\large\ding{194}}.
% ; i.e., we need to determine the frozen blocks and others are searchable blocks as shown in Figure \ref{fig:framework} {\large\ding{194}}.
To address this, the producer conducts 3 steps to determine the frozen blocks for a given backbone architecture.

First, a batch of minority data and majority data are streamed into a pre-trained backbone architecture, and we keep the feature maps in between layers.
The second step compares the feature maps among all groups to obtain the feature variation using  the L2-norm.
Third, we fix a threshold $\mathbb{T}$ by multiplying the maximum variation of all layers and a scaling factor $\gamma$, then search for the foremost layer $\mathbb{L}$ whose feature variation exceeds the threshold $\mathbb{T}$.
This is the splitting point, where all layers before $\mathbb{L}$ belong to the frozen blocks, while the rests (include $\mathbb{L}$) belong to the searchable blocks.

For frozen blocks, in the optimization process, we will directly use the pre-trained parameters (i.e., weights) without training.
Kindly note that in order to reduce the model size to meet the timing constraint, we can further replace the first layers as a convolution layer (which can be trained) and connect them with the frozen blocks to extract common features.
For each searchable block, {\large\ding{192}} controller will determine a set of hyperparameters, including block type, from {\large\ding{193}} search space. After that, the producer will generate a child network $N$ for {\large\ding{195}} evaluator and trainer.
Experimental results will show the effectiveness of the proposed freezing method.

\vspace{1pt}
{\large\ding{195}} \textbf{Evaluator and Trainer}: After a child network is generated, it will be processed by the evaluator and trainer.
The basic design concept is to accelerate the search process.
To achieve this goal, we will first check whether the hardware specification can be met. If not, it will bypass the lengthy training procedure and directly generate the reward as -1 (see Equation \ref{equ:reward}).
To further accelerate the evaluation and enable the automation of the optimization, we will test the performance of each block offline on the given hardware device $H$, based on which we can efficiently estimate the latency during the search process. For the finally identified neural network architecture, we will perform an end-to-end evaluation on the target devices.

If the hardware specification can be met, the searchable blocks in the child network $N$ will be trained to learn a function $f^\prime_N$ for dataset $D$. Then, the trained model will be applied to dataset $D$ and subgroups in $D$ to obtain the model accuracy $A(f_N^{\prime},D)$ and the unfairness score $U(f_N^\prime,D)$.

\section{Experiments}
FaHaNa, which has the demand to be run on mobile phones, is evaluated on a dermatology dataset for diagnosing the dermatological disease. Therefore, we apply two edge devices, Raspberry PI and Odroid XU-4, as our testbed.
Results show that FaHaNa can improve the fairness without compromising accuracy, meanwhile, reducing the model size.

% identify small neural network without losing accuracy but significantly improve the fairness from 15.14\% to 56.36\%.
% Benefiting from the small size, the latency is significantly reduced, achieving more than 10$\times$ reduction over MnasNet and ProxylessNAS, and more than 5$\times$ over MobileNetV2.

% \todo{This section will first (1) introduce the settings of experiments, then report the evaluation results by sets of experiments about (2) FaHaNa exploration in terms of model size, efficiency, and search time, to demonstrate the effectiveness of FaHaNa; (3) comparison with existing approaches for the identified neural network architectures; (4) compatibility of FAHaNa to other fairness technologies.}

\vspace{2pt}
\noindent\textbf{4.1. Experimental Setup}

% The dataset can be divided into two parts. The first part is from 

\textbf{\textit{A. Dataset:}} A dermatology dataset is built based on patients' images collected in the field and the open-access datasets including ISIC 2019 \cite{ISIC2019} for light-skin, Dermnet \cite{Dermnet}, and Atlas dermatology \cite{Altas} for dark-skin.
% The public datasets include ISIC 2019 \todo{[cite]} for light-skin patients, Dermnet\todo{[cite]} and Atlas dermatology\todo{[cite]} for dark-skin patients.
% with 21,587 light-skin images, Dermnet\todo{[cite]} with \todo{} dark-skin images, and Atlas dermatology\todo{[cite]} with 11,255 dark-skin images.
These images are utilized for a classification task with 5 dermatology diseases: Melanoma, Melanocytic nevus, Basal cell carcinoma, Dermatofibroma, and Squamous cell carcinoma.

\textbf{\textit{B. FaHaNa settings:}}
In the evaluation, both parameters $\alpha$ and $\beta$ of the RNN controller (Figure \ref{fig:framework} {\large\ding{192}}) are set to be 1 with the objective to find a neural architecture with balanced accuracy and fairness. 
The {\large\ding{194}} \textbf{Producer} takes MobileNetV2 as the backbone architecture; parameter $\gamma$ is set to 0.5 to select the frozen blocks.
During the search process, we split the dataset into three sets: (1) training set with 60\% images; (2) validation set with 20\% images; and (3) test set with the rest 20\% images.
The number of episodes for reinforcement learning is set to 500.
Finally, a series of neural architectures will be identified, denoted as FaHaNa-Nets.

\textbf{\textit{C. Competitors and training settings:}} To evaluate FaHaNa-Nets, we select a set of state-of-the-art neural networks for comparison, including (1) the manually designed MobileNetV2 \cite{sandler2018mobilenetv2} and ResNet \cite{targ2016resnet}, and (2) the AutoML identified MobileNetV3 \cite{Howard_2019_ICCV}, ProxylessNAS \cite{cai2018proxylessnas} and MnasNet \cite{tan2019mnasnet}.
% \noindent\textbf{\textit{C. Training environment:}} 
For a fair comparison, all the neural networks, including FaHaNa-Nets, are trained from scratch with the same hyperparameters on a GPU cluster with 48 RTX 3080: (1) learning rate starts from 0.1 with a decay of 0.9 in 20 steps, (2) 32 for the batch size, and (3) 500 epochs for training.
In addition, we employ the multi-objective NAS (denoted as MONAS) \cite{hsu2018monas} to {evaluate the efficiency of the FaHaNa framework.}

% , and (4) 500 episode for reinforcement learning. All model training and NAS are executed on a GPU cluster with 48 RTX 3080.

\textbf{\textit{D. Edge devices: }} To compare the inference latency of FaHaNa-Nets and competitors, we employ two kinds of edge devices: (1) Raspberry PI Model B \cite{RaspberryPI} with Broadcom BCM2711 equipping a 1.5 GHz quad-core ARM Cortex-A72 processor and 8 GB memory, and (2) Odroid XU-4 \cite{Odroid} with a Samsung Exynos 5422 equipping ARM Cortex-A15 and Cortex-A7 quad-core processor and 2 GB memory.
% The Raspberry Pi 4 Model B\todo{[cite]} contains a Broadcom BCM2711 with a 1.5 GHz 64-bit quad-core ARM Cortex-A72 processor and 8 GB memory. And the Odroid XU-3\todo{[cite]} which uses a Samsung Exynos 5422 Octa with a 2.0 GHz ARM Cortex-A15 quad-core and a Cortex-A7 quad-core CPUs, 2 GB memory is used. 
The latency is obtained by deploying the trained models on both devices for inference using a vanilla PyTorch framework.

% predict the results of pictures. And the latency of different models on both devices are measured. 

% . The dataset for ISIC 2019 contains 25,331 images available for the classification of dermoscopic images among nine different diagnostic categories: We choose the 5 types among them: Melanoma, Melanocytic nevus, Basal cell carcinoma, Dermatofibroma and Squamous cell carcinoma. It is the majority of the dataset. This part is considered as the white pictures.

% Dermnet is based out of Portsmouth NH with a growing list of contributers from various medical schools and academic institutions which is the largest dermatology source online built to provide online medical education. The categories include acne, melanoma, Bullous disease, Vascular Tumors, etc. Atlas dermatology contains 11255 pictures of dermatology diseases. All images are free to use for non-comercial purposes. We collect the non-white photos from this two websites and considered this combination as the minority part in our project. All this photos are have brown or black skins.

% \clearpage

\vspace{2pt}
\noindent\textbf{4.2 Exploration by FaHaNa}

In the first set of experiments, we demonstrate that FaHaNa-Nets can significantly push forward the Pareto frontiers among fairness, accuracy, and model size, compared with the competitors.
The efficacy of FaHaNa's search engine is also evaluated.

% {Next, we will show the impact of FaHaNa framework on pareto frontiers Figure \ref{fig:FaHaNa} illustartes the extension for the pareto frontier in different angles. Comparing the two exploration results in Figure \ref{fig:FaHaNa}(a) and (b), we can  see that the ideal solution is found by FaHaNa. At the same time, Table \ref{tab:addlabel} gives more information of the comparation results for FaHaNa and Mo-NAS. Detailed analysis will be given out as in the following section.}

\begin{figure}[t]
  \centering
  \includegraphics[width=3.3in]{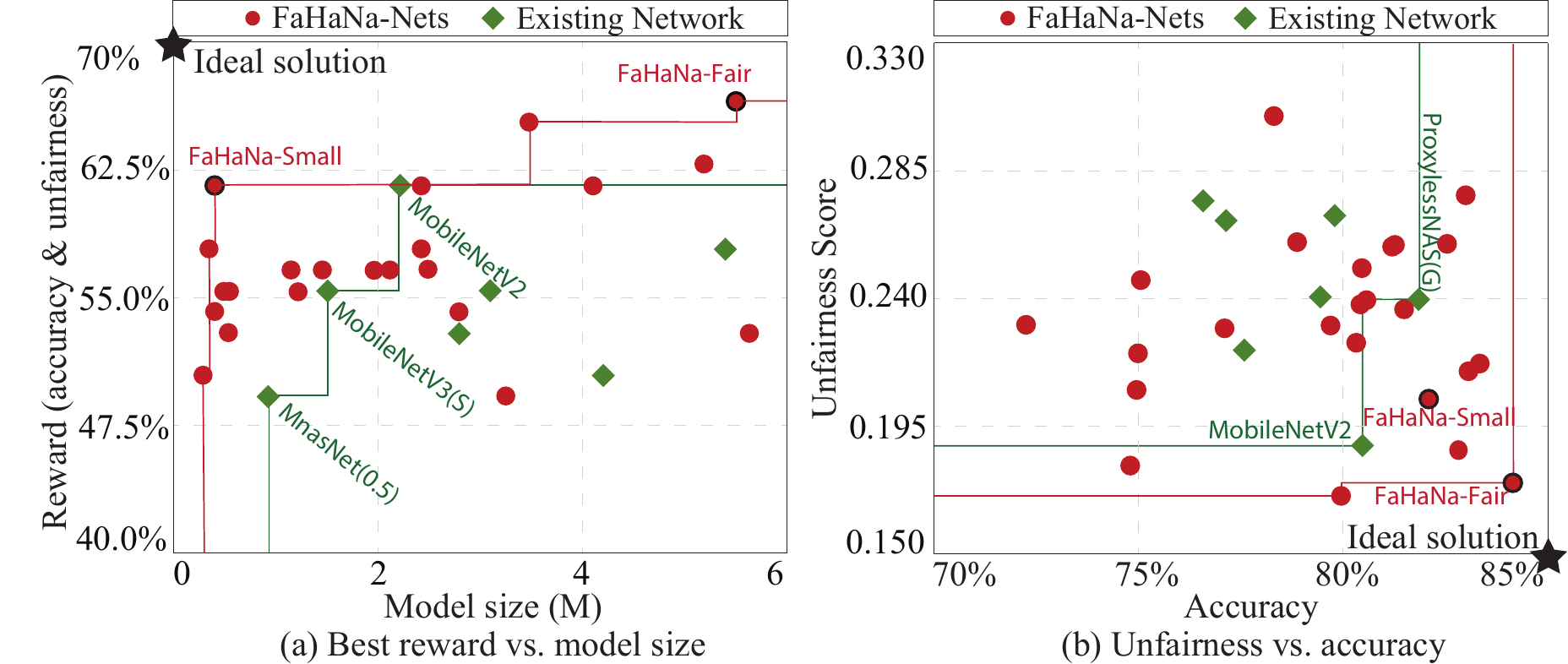}
  \caption{Comparison between the existing neural networks and FaHaNa-Nets with the highest reward.}
  \vspace{-8pt}
  \label{fig:FaHaNa}
\end{figure}

%\vspace{2pt}
\textbf{\textit{A. Best reward vs. model size:}} 
 Figure \ref{fig:FaHaNa} reports the design space exploration results.
% on the tradeoff between the best reward and model size. 
In Figure \ref{fig:FaHaNa}(a), the x-axis is the model size (i.e., number of parameters), and the y-axis is the reward calculated based on Equation \ref{equ:reward}. 
% If the reward is higher, the model is better. 
The ideal solution is located in the left corner, denoted as a star. 
For a clear demonstration, we only plot the architectures with less than 6M parameters. 

In Figure \ref{fig:FaHaNa}(a), each circled point corresponds to a FaHaNa-Net and each green diamond is related to an existing network. 
The red and green lines plot the Pareto frontiers of FaHaNa-Nets and existing networks, respectively.
From this figure, we observe that FaHaNa-Small on the left-top corner dominates all the existing neural networks in terms of reward and model size; while FaHaNa-Fair on the right-top corner achieves the highest fairness.
% Based on these results, we select two FaHaNa-Nets (i.e., FaHaNa-Small and FaHaNa-Fair) for th
% For the green line,it consists of three points which is MnasNet(0.5),MobilenetV3(S) and MobilentV2. MobilentV2 dominates all 
% architectures and gets the highest reward for the existing networks. As for the red line, FaHaNa finds two ideal solution. The first one is the FaHaNa-small whose model size is far less than MobilenetV2 but gets the same reward as it. The second one is the FaHaNa-fair even though the model size is larger than MobilenetV2,it gets the highest reward for the whole architectures including the existing networks and networks found by FaHaNa. 
These figures clearly show that FaHaNa can significantly push forward the Pareto frontiers in the reward and model size tradeoff.
% by further exploring neural architectures. 

%\vspace{2pt}
\textbf{\textit{B. Accuracy vs. Unfairness.}} 
We further investigate the Pareto frontier between fairness and accuracy by decomposing the reward in Figure \ref{fig:FaHaNa}(a).
Results in Figure \ref{fig:FaHaNa}(b) consistently show that FaHaNa can push forward the Pareto frontier compared with the existing neural networks.
More specifically, FaHaNa-Fair is the architecture that is the closest to the ideal solution.
On the other hand, even FaHaNa-small has the smallest size, it can still dominate most of the existing neural architectures.
These two architectures will be used for further detailed comparison.

% Figure \ref{fig:FaHaNa} (b) reports the design space exploration unfairness vs. accuracy. 
% We split the y-axis of the Figure \ref{fig:FaHaNa} (a) to generate the Figure \ref{fig:FaHaNa} (b) to give out more details about the FaHaNa framework. 

% The ideal solution is in the right corner of the picture. Like the Figure \ref{fig:FaHaNa} (a),the red circled points represents the architectures found by FaHaNa and the green diamonds represent the existing network. There are 2 architectures in the green line which are mobilenetV2 and proxylessnas(G). MobilenetV2 gets a lower accuracy but proxylessnas(G) achieves a higher accuracy. As a result, proxylessnas(G) is better than mobilentV2 regarding the accuracy. For the unfairmess,the situation has reversed. MobilenetV2 is much fairer than proxylessnas(G). In general, this two points domain all existing archhitectures. Secondly, we will look at the architectures found by FaHaNa. FaHaNa-Fair is the architecture which is closest to the ideal solution. Considering the model size of FaHaNa-small, even though it is not in the Pareto frontier,the accuracy and unfairness of it are all in the tolerance. 

% Table generated by Excel2LaTeX from sheet 'Sheet3'
\begin{table}[t]
 %\vspace{-5pt}
  \centering
  \small
  \tabcolsep 1.5pt
  \renewcommand\arraystretch{1.0}
  \caption{Effectiveness of freezing method in NAS}
    \begin{tabular}{ccccccccc}
      \hline
          {\multirow{2}{*}{Model}} & \multicolumn{2}{c}{\multirow{2}{*}{Space}} & \multicolumn{3}{c}{Tight Timing Constraint (TC)} & \multicolumn{3}{c}{Relaxed TC} \\
\cline{4-9}          
          & \multicolumn{2}{c}{} & \multicolumn{1}{c}{Valid } & \multicolumn{1}{c}{Time} & \multicolumn{1}{c}{Speedup} & \multicolumn{1}{c}{Valid} & \multicolumn{1}{c}{Time} & \multicolumn{1}{c}{Speedup} \\
\hline
    MONAS & \multicolumn{2}{c}{$10^{19}$}&27.50\%  & 104H45M & 1.00 & 33.33\% & 177H15M & 1.00  \\
    FaHaNa & \multicolumn{2}{c}{$10^{9}$} & 71.05\% & 57H10M  & 1.83 & 95.23\%  & 66H20M  & 2.67  \\
    \hline
    \end{tabular}%
  \label{tab:compare}%
  %\vspace{-5pt}
\end{table}%

% Table generated by Excel2LaTeX from sheet 'To overleaf'
\begin{table*}[t]
   \vspace{-4pt}
  \centering
  \small
  \tabcolsep 1.3pt
  \renewcommand\arraystretch{1.0}
  \caption{Comparison of the existing models and FaHaNa-Nets: Group 1 includes models with limited size (\# of parameters) to be less than 4M and has the accuracy requirement of 81\%; Group 2 includes larger models with the accuracy requirement of 83\%}
    \begin{tabular}{|c|c|c|cc|cc|cc|c||cc|cccc|}
    \hline
    \multirow{2}{*}{Group} & \multirow{2}{*}{ Model} & \multirow{2}{*}{ \# of Para.} & \multirow{2}{*}{ Acc.} &
    { Meet} &
    \multirow{2}{*}{ Light} & \multirow{2}{*}{ Dark}  & Unfairness & { Fairness} &
    \multirow{2}{*}{ Reward} & \multicolumn{2}{c|}{ Storage} & 
    \multicolumn{4}{c|}{ Latency (ms)}   \\ 
    &&&& Acc. &&& Score & Comp. && { (MB)} & Red. & { Raspberry} & { Speedup} 
    & { Odroid} & { Speedup} \\
    \hline
          & MobileNetV2 & 2,230,277 & 81.05\% & \checkmark & 81.27\% & 58.02\% & 0.2325  & baseline & 0.58 & 8.51  & baseline  & 1939.40 & baseline & 4264.55 & baseline  \\
          & ProxylessNAS(M) & 2,805,917 & 81.27\% & \checkmark & 81.56\% & 50.62\% & 0.3094   & 33.08\% $\downarrow$ & 0.50 & 10.70  & 0.79$\times$ & 5241.51 & 0.37$\times$  & 8784.53 & 0.49$\times$ \\
          G1 & MnasNet 0.5 & 943,917 & 78.12\% & $\times$ & 78.54\% & 33.33\% & 0.4521  & 94.45\% $\downarrow$ & -1.00 & 3.60  & 2.36$\times$  & 714.19 & 2.72$\times$ & 2312.05 & 1.84$\times$ \\
          ($<4M$)& MobileNetV3(S) & 1,522,981 & 80.38\% & $\times$ & 80.68\% & 48.15\% & 0.3253   & 39.91\% $\downarrow$ & -1.00 & 5.81  & 1.46$\times$ & 658.84 & 2.94$\times$ & 1954.14 & 2.18$\times$ \\
          & MnasNet 1.0 & 3,108,717 & 80.71\% & $\times$ & 80.98\% & 51.85\% & 0.2913   & 25.29\% $\downarrow$ & -1.00 & 11.86  & 0.72$\times$ & 3855.72 & 0.50$\times$ & 7033.29 & 0.61$\times$ \\
          & \textbf{FaHaNa-Small} & \textbf{422,341} &\textbf{ 81.28\%} &\textbf{ \checkmark }& \textbf{81.46\% }&\textbf{ 61.73\%} & \textbf{0.1973 } & \textbf{15.14\% $\uparrow$} &\textbf{0.62} & \textbf{1.61 } & \textbf{5.28$\times$} & \textbf{337.30} & \textbf{5.75$\times$} & \textbf{736.22} & \textbf{5.79$\times$} \\

    % \multirow{6}{*}{$<4M$} & MobileNetV2 & 2,230,277 & 81.05\% & 0.00\%     & 81.27\% & 58.02\% & 23.25\% & 0.00\%     & 57.80  & 8.51  & 1939.4 & 1.00$\times$     & 4264.55 & 1.00$\times$ \\
    % & ProxylessNAS(M) & 2,805,917 & 81.27\% & 0.22\% & 81.56\% & 50.62\% & 30.94\% & -7.69\% & 50.33  & 10.70  & 5241.51 & 0.37$\times$  & 8784.53 & 0.49$\times$  \\
    % & MnasNet 0.5 & 943,917 & 78.12\% & -2.93\% & 78.54\% & 33.33\% & 45.21\% & -21.96\% & 32.91  & 3.60  & 714.19 & 2.72$\times$  & 2312.05 & 1.84$\times$  \\
    % & MobileNetV3(S) & 1,522,981 & 80.38\% & -0.67\% & 80.68\% & 48.15\% & 32.53\% & -9.28\% & 47.85  & 5.81  & 658.84 & 2.94$\times$  & 1954.14 & 2.18$\times$  \\
    % & MnasNet 1.0 & 3,108,717 & 80.71\% & -0.34\% & 80.98\% & 51.85\% & 29.13\% & -5.88\% & 51.58  & 11.86  & 3855.72 & 0.50$\times$  & 7033.29 & 0.61$\times$  \\
    
    % & \textbf{FaHaNa-Small} & \textbf{422,341} & 81.28\% & 0.23\% & 81.46\% & 61.73\% & \textbf{19.73\%} & \textbf{3.52\%} & 61.55  & \textbf{1.61}  & \textbf{337.3} & \textbf{5.75$\times$}  & \textbf{736.22} & \textbf{5.79$\times$}  \\

    \hline 
    
    \hline 
    
          & ResNet-50 & 23,518,277 & 83.81\% & \checkmark & 83.98\% & 65.43\% & 0.1855  & baseline & 0.65 & 89.72  & baseline  & 1063.61 & baseline  & 5750.42 & baseline  \\
          & ResNet-18 & 11,179,077 & 83.08\% & \checkmark & 83.28\% & 61.73\% & 0.2155  & 16.17\% $\downarrow$ & 0.62 & 42.64  & 2.10$\times$  & 425.90 & 2.50$\times$  & 1373.16 & 4.19$\times$  \\
          G2& ResNet-34 & 21,287,237 & 83.01\% & \checkmark & 83.23\% & 59.26\% & 0.2397  & 29.22\% $\downarrow$ & 0.59 & 81.20  & 1.10$\times$  & 621.87 & 1.71$\times$  & 2829.22 & 2.03$\times$  \\
          ($\ge4M$)& ProxylessNAS(G) & 5,399,493 & 83.21\% & \checkmark & 83.46\% & 56.79\% & 0.2667  & 43.77\% $\downarrow$ & 0.57 & 20.60  & 4.36$\times$  & 3714.44 & 0.29$\times$  & 9426.17 & 0.61$\times$  \\
          & MobileNetV3(L) & 4,208,437 & 79.58\% & $\times$ & 80.00\% & 34.57\% & 0.4543  & 144.91\% $\downarrow$ & -1.00 & 16.05  & 5.59$\times$  & 2668.00  & 0.40$\times$  & 4824.40 & 1.19$\times$  \\
          &\textbf{ FaHaNa-Fair} &\textbf{ 5,502,469} & \textbf{84.06\%} &\textbf{ \checkmark} & \textbf{84.22\%} & \textbf{66.67\% }& \textbf{0.1755} & \textbf{5.39\%$\uparrow$} & \textbf{0.67} & \textbf{20.99 } & \textbf{4.27$\times$ } &\textbf{ 606.80} & \textbf{1.75$\times$ } & \textbf{1833.76} & \textbf{3.14$\times$ } \\

    \hline
    \end{tabular}%
  \label{tab:exp1}%
\end{table*}%

%Lei: 
%Table 2 shows the Effectiveness of the Freezing Approach: 1. FaHaNa shrink the exploration space compared with MO-NAS; 2. Speedup; 3. valid models are far more than MO-NAS
%Figure 5: 

%\vspace{2pt}
\textbf{\textit{C. Space and time.}} 
The efficiency and effectiveness of the freezing method are evaluated by comparing MONAS (with fairness added as one objective).
We compare the search space and search time in Table \ref{tab:compare}.
Two sets of experiments are carried out using a tight timing constraint and a relaxed timing constraint.
Columns ``Valid'' show the ratio of the valid architectures (i.e., the reward is not equal to $-1$, see Equation \ref{equ:reward}) examined during NAS process.

There are several observations in Table \ref{tab:compare}.
First,
% reports the comparison results on 
% MO-NAS \todo{[todo]} and FaHaNa. Search space is Significantly reducing from {$10^{19}$} to {$10^{9}$}. 
FaHaNa can significantly reduce the search space, from {$10^{19}$} to {$10^{9}$}, compared with MONAS.
Second, benefiting from the reduced search space, FaHaNa can search for more valid architectures.
With the same number of episodes, the validation rates of MONAS and FaHaNa are increased from 27.50\% to 71.05\% and from 33.33\% to 95.23\% under tight TC and relaxed TC, respectively.
This is because the freezing method can prune a lot of invalid neural architectures.
Third, even with a higher validation rate (more architectures need to be trained), FaHaNa can still achieve 1.83$\times$ and 2.67$\times$ speedup.% against MONAS.
This is because the freezing method can reduce the number of parameters to accelerate the network training process.
Overall, FaHaNa can shrink the search space to exam more valid networks for high-reward architectures; meanwhile, the search time can be significantly reduced.

\vspace{2pt}
\noindent\textbf{4.3 FaHaNa-Nets vs. Existing Neural Architectures}

Next, we compare FaHaNa-Nets against competitors with a given accuracy constraint (AC).
We divide all neural architectures into two groups in terms of model size.
Group G1 contains the small-size architectures with less than 4M parameters; other architectures belong to group G2.
We select the architecture with the highest fairness from all the competitors in each group as the baseline: MobileNetV2 for G1 and ResNet-50 for G2.
Table \ref{tab:exp1} reports the fairest trained model for each architecture, which is expected to meet a preset AC: 81\% for G1 and 83\% for G2.
If the architecture cannot meet AC, then we select the model with the highest accuracy for comparison.
The parallel lines divide Table \ref{tab:exp1} into two parts: software metrics (left) and hardware metrics (right).
% Columns  on the left of the parallel lines report the software related metrics, including number of parameters, model accuracy, accuracy for minority and majority groups, unfairness scores and reward.
% We also report the hardware related metrics, including storage requirements and latency.

% The baseline for group 1 is the MobileNetV2 which performs best among all competitors in terms of fairness; while the baseline for 
% Group 1 has 6 architectures, whose parameters is less than 4 millions in total; and group 2 also has 6 archite
% If its total parameters  is larger than 4 millions, it will in the group 2 otherwise it will in the group 1. So we will analyse the Table \ref{tab:exp1} in 2 grops.

%\vspace{2pt}
\textbf{\textit{A. FaHaNa-Small has the smallest size and lowest latency:}}
From Table \ref{tab:exp1}, we have several observations.
First, only MobileNetV2, 
% For this constraint, only mobilenetV2 and 
ProxylessNAS(M), and FaHaNa-Small meet the AC of 81\%.
% achieves the target and other existing networks all eliminate. 
% As for the unfaireness, MobileNetV2 gets the lowest value,it means mobilenetv2 is much fairer than proxylessnas(M) so this architecture is selected as the baseline for this group. 
% Every model's unfairness will be compared with mobilenetv2. 
Second, FaHaNa-Small is the fairest architecture in G1. Compared with the baseline, MobileNetV2 with a
{0.2325}
%23.25\% 
unfairness score, FaHaNa-Small can get {0.1973}
%19.73\%
which has a {15.14\%}
%3.52\% 
{improvement}.
Compared with other architectures, the fairness improvement of FaHaNa-Small can reach up to {56.34\% (i.e., MnasNet 0.5).}
Third, FaHaNa-Small has the minimum number of parameters; thus, it has the best hardware performance: 1.61M storage, 337.3ms latency on Raspberry PI, and 736.22ms latency on Odriod.
Compared with the baseline, it achieves 5.28$\times$ storage reduction, as well as 5.75$\times$ and 5.79$\times$ speedup on Raspberry PI and Odriod respectively.
These results, in response to our initial question, verified we can find a small neural network to achieve fairness for edge devices.

% For the storage which is calculated from the parameters, FaHaNa-Small reduce 5.28$\times$. 
% In the last step,we put every architecture in the different hardware platform, Raspberry PI and Odroid, to test the latency. In the Raspberry PI, the latency for the mobilenetv2 is 1939.4ms but for the FaHaNa-Small ,it is just 337.3ms. It achieves 5.75$\times$ speed up. For the Odroid,mobilenetV2 gets 4264.55ms and FaHaNa-Small only need 736.22ms. It gets 5.79$\times$ speed up.

% \vspace{2pt}
% \textbf{\textit{A. Visualization of Models from FaHaNa.}} 
% \vspace{2pt}

% \begin{figure*}[t]
%   \centering
%   \includegraphics[width=7in]{figure/Exp3.eps}
%   \caption{Architecture one found by NAS}
%   \label{fig:framework}
% \end{figure*}

\begin{figure}[t]
  \centering
%   \vspace{5pt}
  \includegraphics[width=3.1in]{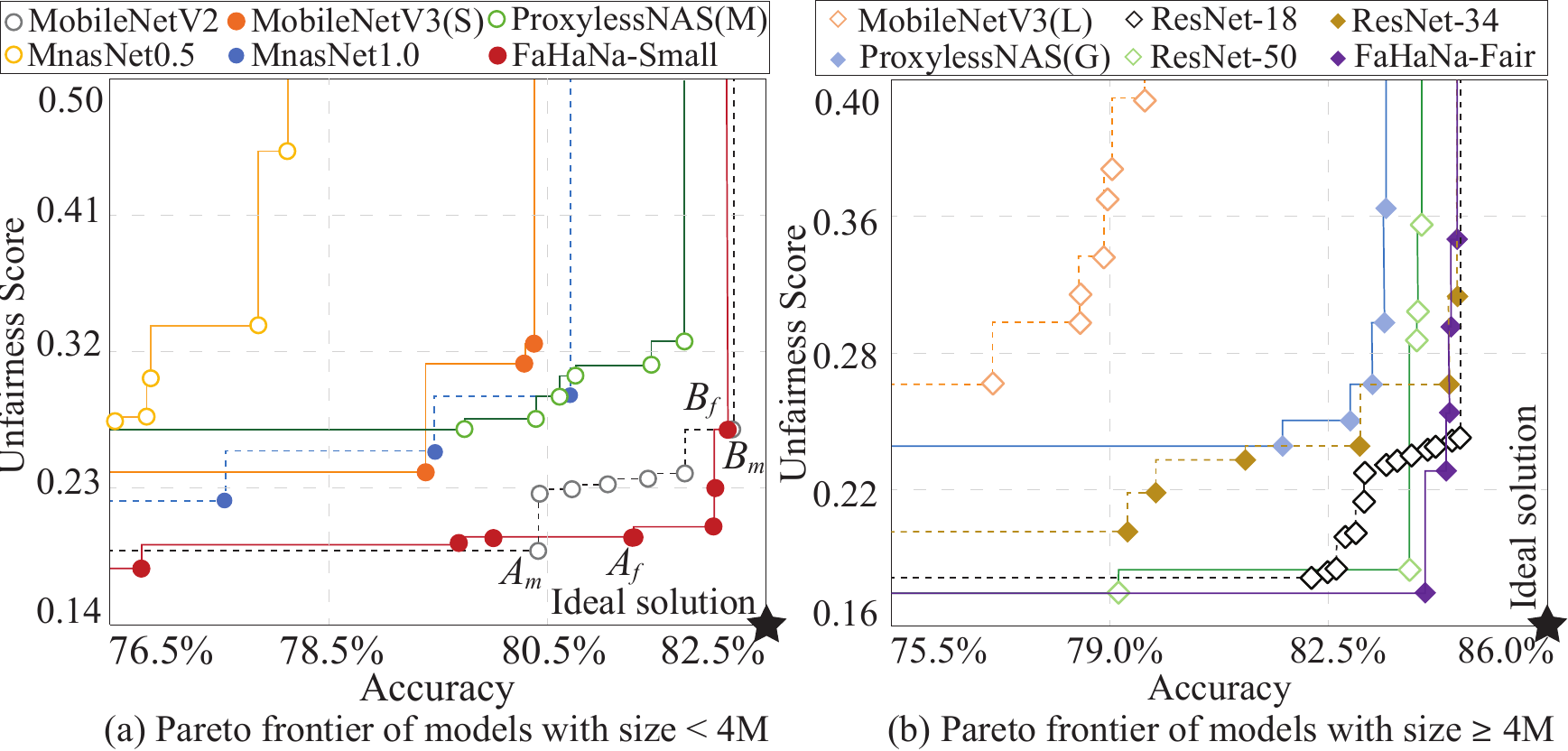}
  \caption{Pareto frontiers of the existing models, FaHaNa-Small and FaHaNa-Fair in terms of accuracy and unfairness.}
    \vspace{-8pt}
  \label{fig:Paretofrontiers}
\end{figure}

%\vspace{2pt}
\textbf{\textit{B. FaHaNa-Fair can achieve the highest fairness:}}
FaHaNa-Fair is the fairest model in all competitors.
Similar to the results in G1, FaHaNa-Fair achieves the lowest unfairness score in G2,
{0.1755}, compared with {0.1855}
%17.55\%%18.55\% 
obtained by the baseline architecture ResNet-50. In addition, FaHaNa-Fair is 4.27$\times$ smaller than ResNet-50, achieving 1.75$\times$ and 3.14$\times$ speedup on edge devices.
\textbf{\textit{C. Pareto frontier:}}
Figure \ref{fig:Paretofrontiers} further shows the comparison of Pareto frontiers in terms of the accuracy-unfairness tradeoff built by all models. 
% The x-axis and y-axis represent the accuracy and unfairness, respectively.
Figure \ref{fig:Paretofrontiers}(a) and Figure \ref{fig:Paretofrontiers}(b) show the results of the models in G1 and G2, respectively.
% while the figure \ref{fig:framework} (b) shows the result of the modes with size bigger than 4M. 
The stars in these figures refer to the ideal solutions. 
In Figure \ref{fig:Paretofrontiers}(a), the red points form the Pareto frontier of FaHaNa-Small.
It is clear that FaHaNa-Small dominates all other competitors except MobileNetV2.
In the comparison with MobileNetV2, FaHaNa-Small performs better in most cases.
There are only two special cases, $A_m$ and $B_m$ from MobileNetV2, escaping the domination of FaHaNa-Small.
But, the unfairness {score} gap between $A_m$ and $A_f$ is only 0.0084, while the accuracy gap between $B_m$ and $B_f$ is 0.04\%.
Similarly, FaHaNa-Fair dominates almost all architectures in group G2.
All these results show the superiority of FaHaNa-Nets over the existing small-size neural architectures.

\begin{table}[t]
  \centering
    \small
  \tabcolsep 3.8pt
  \renewcommand\arraystretch{1.0}
  \caption{FaHaNa-Nets can consistently achieve better fairness when data balancing \cite{choi2020fair} is applied for improving fairness}
    \begin{tabular}{|c|cc|cccc|}
    \hline
        \multirow{2}{*}{Model}  & \multicolumn{2}{c|}{w/o balancing} & \multicolumn{4}{c|}{w/ balancing} \\
    \cline{2-7}
     & Acc. & Unfair. & Acc. & Impr. & Unfair. & Impr. \\
    \hline
    MobileNetV2 & 81.05\% & 0.2325 & 82.14\% & 1.09\% & 0.1528 & 0.0797 \\
    
    ProxylessNAS(M) & 81.27\% & 0.3094 & 81.53\% & 0.26\% & 0.1467 & 0.1627 \\
    MnasNet 0.5 & 78.12\% & 0.4521 & 78.82\% & 0.70\% & 0.1824 & 0.2697  \\
    MobileNetV3(S) & 80.38\% & 0.3253 & 80.55\% & 0.17\% & 0.1923 & 0.1330 \\
    MnasNet 1.0 & 80.71\% & 0.2913 & 80.20\% & -0.51\% & 0.1585 & 0.1328  \\
    FaHaNa-Small & 81.28\% & 0.1973 & 82.02\% & 0.74\% & 0.1365 & 0.0608 \\
    \hline
    \end{tabular}%
  \label{tab:exp4}%
  %\vspace{-5pt}
\end{table}%

\vspace{2pt}
\noindent\textbf{4.4 Compatibility of FaHaNa with Data Balancing Techniques}

One typical approach for fairness improvement is to 
% Existing works apply generative models to 
generate more minority data
% for improving fairness 
\cite{choi2020fair}. 
In Table \ref{tab:exp4}, we show the proposed FaHaNa framework is compatible with the data balancing techniques.
We apply the same method in \cite{choi2020fair} to get $5\times$ more minority data for training.
% to train the existing networks and FaHaNa-Small. We set 500\% for data growth rate.  
It is obvious that after data balancing, all networks except MnasNet 1.0 can improve both accuracy and fairness; even for MnasNet 1.0, it can achieve a {0.1328} lower unfairness score in fairness with 0.51\% accuracy degradation.
From the results in Table \ref{tab:exp4}, FaHaNa-Small can also get benefits from data balancing to improve accuracy by 0.74\% while achieving {0.0608}  fairness improvement.
What's more, FaHaNa-Small is still the fairest model.
% except MnasNet 1.0, all networks become much fairer compared with no data balancing. Without data balancing,mobilenetV2 is the fairnest model in the existing networks,however after the data balancing,proxylessNAS(M) becomes the fairest model. The accuracy for the proxylessNAS(M) is increasing from 81.27\% to 81.53\%.
% For FaHaNa-Small, it is also sensitive to the data balancing, the accuracy increases to 82.02\%. The improvement for the FaHaNa-Small is 0.74\% and there is only 0.26\% improvement for proxylessnas(M).  Then we will see the unfairness improvement. For the proxylessnas(M), the unfairness reaches 14.67\%. As for the FaHaNa-Small, the unfairness achieves 13.65\%. It proves that FaHaNa-Small is also sensitive to the data balacing. And it 

%MobileNetV2 gains 1.09\% improvement for the accuracy which is the highest improvement rate. The accuracy increases from 81.05\% to 82.14\%. For proxylessNAS(M),MnasNet 0.5 and mobilenetV3(S),the improvements are all less than 1\%. 

% \clearpage

\vspace{2pt}
\noindent\textbf{4.5 Insights from FaHaNa-Nets}

\begin{figure}[t]
  \centering
  \includegraphics[width=3in]{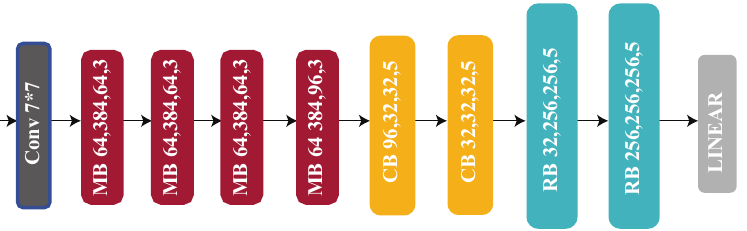}
  \caption{Visualization of FaHaNa-Fair.}
   \vspace{-2pt}
  \label{fig:visualres}
  \vspace{-2pt}
\end{figure}

Figure \ref{fig:visualres} provides the visualization of FaHaNa-Fair.
An insightful observation is that we applied MB block to extract common features in the head layers while utilizing larger blocks (e.g., CB and RB) at the end layers to address the fairness issue.
Such an architecture can make a good tradeoff between hardware specifications and fairness requirements: (1) the head layers with high resolution apply MB block for fewer parameters; and (2) the end layers are sensitive to fairness thus CB and RB blocks are applied to achieve higher fairness.
The insight here is that a homogeneous design with the same type of block can not balance accuracy, fairness, and latency, but FaHaNa can due to its flexibility in block selection.

\vspace{-5pt}

\section{conclusion}
\label{sec:conclusion}

In this work, we have proposed a fairness- and hardware-aware NAS framework, FaHaNa, integrating fairness in NAS for the first time to design the fair neural architecture. On top of it, a freezing method has been proposed to accelerate the NAS process.
%In this work, we identify the problem that the neural architecture can affect the fairness, but how to design a fair neural architecture is still unclear; no more saying the identification of fair architecture with hardware constraints. To address these issues, a fairness- and hardware-aware NAS framework, namely FaHaNa, was proposed. We not only integrated fairness in NAS for the first time, but also developed a freezing method to accelerate the NAS process.
As such, FaHaNa can identify a series of neural architectures forming a much better Pareto frontier on accuracy, fairness, and model size, compared to the existing neural architectures.
Moreover, FaHaNa is compatible with the existing techniques for fairness improvement.
Extensive experiments are carried out to evaluate FaHaNa, where architecture with $5\times$ smaller size and $5\times$ lower latency can be obtained for edge devices, meanwhile, achieving 15.14\% higher fairness and not compromising overall accuracy, compared to MobileNetV2 which has the highest fairness in all examined competitors.

% We proposed FaHaNa framework to speed up neural architecture search. Most importantly, we find FaHaNa-Fair to solve the fairness issue. This is driven by the trend that
% more and more networks are deployed in the edge devices. This paper took the FaHaNa framework as a vehicle to show that through exploring decreased search space, 
% tradeoffs can be significantly pushed forward. 
% FaHaNa framework will be the base for fair neural architecture search. We list two promising architectures FaHaNa-Small and FaHaNa-Fair. 
% Finally, FaHaNa-Fair and FaHaNa-Small all achieve great success for the latency in the edge device.

\vspace{-2pt}
\bibliographystyle{unsrt2authabbrvpp}
\bibliography{paper}

\end{document}